\documentclass[11pt]{article}

\usepackage[preprint]{acl}
\usepackage{booktabs}
\usepackage{enumitem}
\usepackage{amsmath}
\usepackage{amssymb}
\usepackage{amsfonts}
\usepackage{multirow}
\usepackage{placeins}

\usepackage{times}
\usepackage{latexsym}

\usepackage[T1]{fontenc}

\usepackage[utf8]{inputenc}

\usepackage{microtype}

\usepackage{inconsolata}

\usepackage{graphicx}

\title{SenFlow: Inter-Sentence Flow Modeling for AI-Generated Text Detection in Hybrid Documents}

\author{
	Jingkun Luo\textsuperscript{1} \quad
	Yifan Sun\textsuperscript{1} \quad
	Da-Tian Peng\textsuperscript{1}\thanks{\,Corresponding author.} \quad
	Guanxiong Pei\textsuperscript{2} \\
	\textsuperscript{1}Northwestern Polytechnical University \quad
	\textsuperscript{2}Zhejiang Lab \\
	\texttt{\{luojingkun, sunyf\}@mail.nwpu.edu.cn} \quad
	\texttt{pengdatian@nwpu.edu.cn} \quad
	\texttt{pgx@zhejianglab.org}
}

\begin{document}
	\maketitle
	\begin{abstract}
		Sentence-level AI-generated text detection (S-AGTD) for hybrid documents, where humans and LLMs co-author one text, faces two gaps: existing methods classify each sentence in isolation, discarding inter-sentence dependencies, and existing benchmarks omit the newest generation of generators. We construct \textbf{MOSAIC}, a benchmark of 16{,}000 hybrid documents over PubMed and XSum, generated by DeepSeek-V3.2 and Kimi K2 under stringent quality controls including a perplexity-consistency filter absent from prior benchmarks. We recast S-AGTD as structured prediction over the document sentence sequence and instantiate it as \textbf{SenFlow}, integrating graph-based inter-sentence propagation with linear-chain CRF decoding in a single document-level pass over a sentence graph. SenFlow reaches state-of-the-art performance on MOSAIC, with a +4.15 pp average Macro-F1 margin on cross-domain transfer, the hardest of three protocols of increasing difficulty. We further find that even after the perplexity filter equalizes overt cues, AI insertions retain a generator-dependent sentence-length gap that sentence-level detectors still exploit. Code and data: \url{https://github.com/luojingkun22/SenFlow}.
	\end{abstract}
	
	\section{Introduction}
	\label{sec:intro}
	
	\begin{figure}[!t]
		\centering
		\includegraphics[width=\columnwidth]{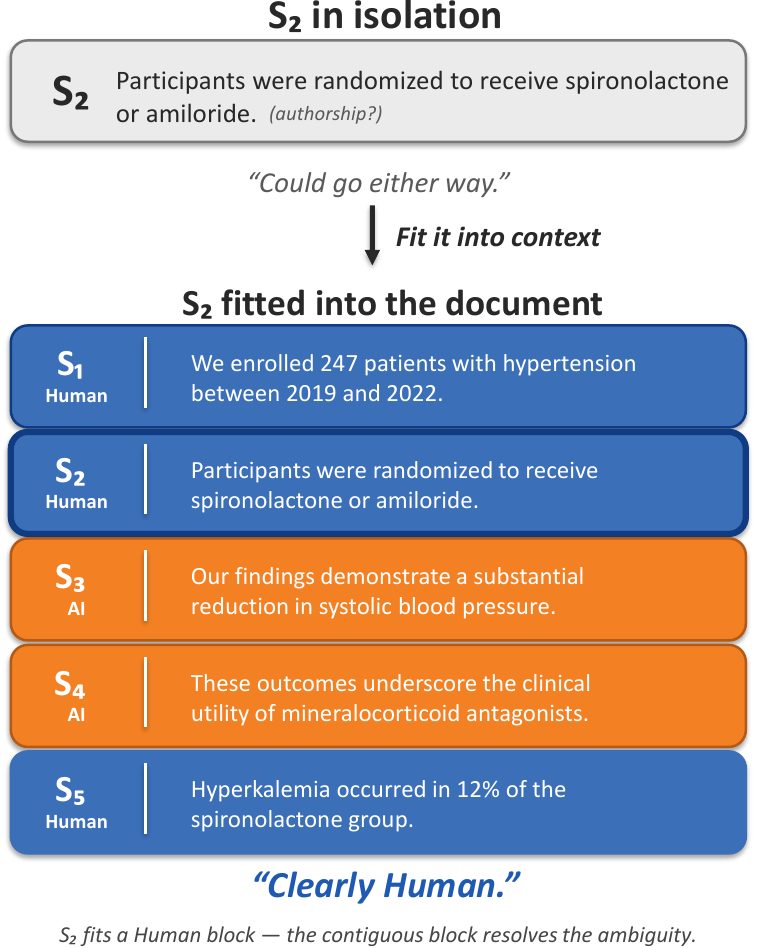}
		\caption{\textbf{S-AGTD is hard in isolation but tractable with context.} Sentence $s_2$ is ambiguous alone; its surrounding context reveals authorship through the local clustering of AI-generated sentences.}
		\label{fig:motivation}
	\end{figure}
	
	The rapid advancement of large language models (LLMs) has made machine-generated text increasingly difficult to distinguish from human writing, raising concerns across academia, journalism, and professional document drafting~\citep{zhao2023survey, wu2025survey}. While document-level AI-generated text detection (AGTD) has advanced rapidly~\citep{mitchell2023detectgpt, bao2024fast, yang2024dna, kirchenbauer2023watermark}, these methods cannot handle \textit{hybrid documents}, in which human-written and machine-generated sentences are interleaved within a single piece of text. Hybrid writing is increasingly prevalent in practice~\citep{lee2022coauthor, zhang2024llm, su2025haco}: a student may polish paragraphs, a researcher may insert AI-generated summaries, or a legal analyst may draft clauses with an LLM. These scenarios require detection at the \textbf{sentence level}, a substantially harder task than document-level classification, since individual sentences offer limited stylistic cues~\citep{sadasivan2023can,weber2023testing}, as illustrated in Figure~\ref{fig:motivation}. As shown by \citet{jiang2025sendetex}, training-free detectors such as Fast-DetectGPT~\citep{bao2024fast} suffer significant degradation at the sentence granularity, with Macro-F1 dropping below 0.85 even on in-domain data.
	
	The sentence-level AGTD (S-AGTD) challenge has attracted growing attention. SeqXGPT~\citep{wang2023seqxgpt} formulates the task as word-level sequence labeling, aggregating predictions to sentence labels via majority vote. SenDetEX~\citep{jiang2025sendetex} introduces a style-context fusion framework that combines token probability and entropy signals with regenerated contextual embeddings. While advancing the state of the art, these methods share a fundamental limitation: each sentence is classified in isolation, without modeling dependencies between neighboring sentences. In hybrid documents, however, authorship transitions are rarely random: AI-generated passages tend to cluster in contiguous blocks, and stylistic contrast typically evolves gradually across sentence boundaries rather than shifting at a single point. Ignoring these inter-sentence patterns discards a structural signal informative for detection.
	
	Beyond methodology, a parallel gap exists in evaluation. Current S-AGTD benchmarks are built on a previous generation of generators, the standard chat models such as GPT-4o and DeepSeek-V3~\citep{jiang2025sendetex}. The generator landscape has since shifted: the newest systems, including reasoning models such as DeepSeek-R1~\citep{guo2025deepseek} and OpenAI o1/o3~\citep{jaech2024openai}, are absent from these benchmarks. Whether S-AGTD methods tuned on earlier chat-model text generalize to current-generation outputs is therefore untested, a coverage gap that existing benchmarks cannot close.
	
	We address both gaps with two complementary artifacts. \textbf{SenFlow} treats each document as a sentence graph, propagates contextual information across sentences via a GCN, and enforces label coherence via CRF decoding. \textbf{MOSAIC} pairs the reasoning model DeepSeek-V3.2 with the chat model Kimi K2 across PubMed and XSum, producing 16{,}000 hybrid documents through block-level masking with stringent quality controls.
	
	\begin{table*}[!tbp]
		\centering
		\resizebox{\textwidth}{!}{%
			\begin{tabular}{@{}ll ll r rrr rc rr@{}}
				\toprule
				\textbf{Subset} & \textbf{Domain} & \textbf{Generator} & \textbf{Type} & \textbf{\#Docs} & \textbf{\#Sents} & \textbf{\#Human} & \textbf{\#AI} & $\bar{N}$ & \textbf{Mask (\%)} & \textbf{PPL\textsubscript{ori}} & \textbf{PPL\textsubscript{hyb}} \\
				\midrule
				PD & PubMed & DeepSeek-V3.2 & Reasoning & \phantom{0}4{,}000 & \phantom{0}79{,}926 & \phantom{0}55{,}951 & 23{,}975 & 20.0 & 30.0 & 8.45 & 8.69 \\
				PK & PubMed & Kimi K2       & Chat      & \phantom{0}4{,}000 & \phantom{0}79{,}923 & \phantom{0}55{,}950 & 23{,}973 & 20.0 & 30.0 & 8.37 & 8.12 \\
				XD & XSum   & DeepSeek-V3.2 & Reasoning & \phantom{0}4{,}000 & \phantom{0}60{,}174 & \phantom{0}42{,}242 & 17{,}932 & 15.0 & 29.7 & 9.55 & 9.83 \\
				XK & XSum   & Kimi K2       & Chat      & \phantom{0}4{,}000 & \phantom{0}60{,}972 & \phantom{0}42{,}790 & 18{,}182 & 15.2 & 29.8 & 9.51 & 8.17 \\
				\midrule
				\multicolumn{4}{@{}l}{Total} & 16{,}000 & 280{,}995 & 196{,}933 & 84{,}062 & \multicolumn{1}{c}{$-$} & \multicolumn{1}{c}{$-$} & \multicolumn{1}{c}{$-$} & \multicolumn{1}{c}{$-$} \\
				\bottomrule
			\end{tabular}%
		}
		\caption{Statistics of MOSAIC. $\bar{N}$ = mean sentences per document. Mask~(\%) = mean proportion of AI-generated sentences. PPL\textsubscript{ori} and PPL\textsubscript{hyb} = mean perplexity of original and hybrid documents measured by Llama-3.1-8B-Instruct. Each subset is split 2{,}800/600/600 by document ID into train, validation, and test. The subset abbreviations PD, PK, XD, XK are used throughout the paper.}
		\label{tab:dataset}
	\end{table*}
	
	Our contributions are fourfold: \textbf{(1)} \textbf{MOSAIC}, a sentence-level S-AGTD benchmark of 16{,}000 hybrid documents pairing reasoning- and chat-model generators across biomedical and news domains, constructed with a perplexity-consistency filter absent from prior S-AGTD benchmarks; \textbf{(2)} an empirical finding enabled by MOSAIC's perplexity-consistency filter: once overt perplexity cues are equalized, AI insertions still retain a generator-dependent sentence-length structural gap against the surrounding human prose, larger for DeepSeek-V3.2 than for Kimi K2, that remains exploitable at the sentence level, evidence that filtering away perplexity-level naturalness does not render insertions undetectable; \textbf{(3)} \textbf{SenFlow}, a structured-prediction formulation of S-AGTD that integrates graph-based inter-sentence propagation with CRF decoding into a single document-level pass over a sentence graph; \textbf{(4)} state-of-the-art results across three protocols of increasing difficulty, with the largest 4.15 pp Avg F1 margin on cross-domain transfer, attained using approximately $20\times$ fewer proxy-model inference calls than the previous best method.

	\section{Related Work}
	
	\textbf{Document-level AGTD.} Document-level methods follow training-free or training-based paradigms. Training-free methods exploit statistical properties of model outputs via log-probability curvature~\citep{mitchell2023detectgpt, bao2024fast}, $n$-gram divergence~\citep{yang2024dna}, paired-LLM contrasts~\citep{hans2024spotting}, and rewrite-asymmetry signals~\citep{mao2024raidar}. Training-based methods include fine-tuned classifiers~\citep{solaiman2019release}, adversarial training~\citep{hu2023radar}, perturbation-resistant reconstruction~\citep{huang2024ai}, and surrogate-target alignment~\citep{zeng2024dald}. Watermarking~\citep{kirchenbauer2023watermark, zhao2023provable} embeds detection signals during generation but requires control over the generator. Out-of-distribution evaluations show detection degrades under unseen domains and generators~\citep{li2024mage}. All assume a single-source document and cannot handle hybrid content.
	
	\textbf{Sentence-level AGTD.} SeqXGPT~\citep{wang2023seqxgpt} pioneered S-AGTD as word-level sequence labeling over token log probabilities. POGER~\citep{shi2024ten} estimates generation probabilities through proxy-guided resampling under black-box settings. The current state of the art is SenDetEX~\citep{jiang2025sendetex}, which combines proxy-model probability and entropy features with regenerated contextual embeddings, and contributes an AutoFill-Refine benchmark using GPT-4o and DeepSeek-V3. Related work measures sentence-regeneration similarity~\citep{nguyen2024simllm} or extends the task to word-level co-authoring attribution~\citep{su2025haco, zhang2024llm}. The latter line operates at a finer, word-level granularity on its own word-tagged benchmarks and is not directly comparable without re-annotating MOSAIC at the word level.
	
	\section{The MOSAIC Benchmark}
	\label{sec:bench}
	
	\begin{figure*}[h]
		\centering
		\includegraphics[width=\textwidth]{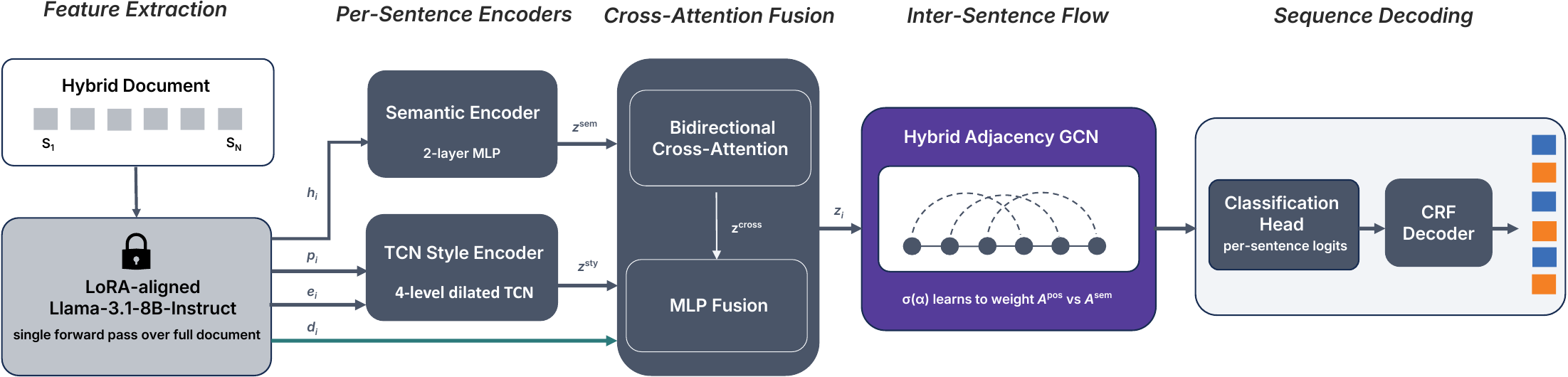}
		\caption{\textbf{SenFlow architecture.} Four per-sentence streams from a frozen LoRA-aligned Llama-3.1-8B-Instruct are fused by dual cross-attention, refined by the Hybrid Adjacency GCN, and decoded by a linear-chain CRF. Solid edges denote positional adjacency $\mathbf{A}^{\text{pos}}$, dashed edges semantic $\mathbf{A}^{\text{sem}}$; blue and amber mark human and AI predictions.}
		\label{fig:overview}
	\end{figure*}
	
	\subsection{Task Definition}
	
	Given a hybrid document $D = (s_1, \dots, s_N)$ of $N$ sentences with mixed authorship, S-AGTD predicts a label sequence $\mathbf{y} = (y_1, \dots, y_N)$ with $y_i \in \{0, 1\}$, where $y_i = 0$ marks a human-written sentence and $y_i = 1$ a machine-generated one; the task requires fine-grained identification of authorship transitions within a single document.
	
	\subsection{Hybrid Corpus Construction}
	
	We construct MOSAIC from two source corpora, biomedical abstracts from PubMed~\citep{cohan2018discourse} and news articles from XSum~\citep{narayan2018don}, using two generators that represent distinct paradigms: \textbf{DeepSeek-V3.2} in thinking mode (\texttt{deepseek-reasoner} API), a reasoning model that performs internal chain-of-thought deliberation before producing outputs, and \textbf{Kimi K2} (\texttt{kimi-k2-0905-preview} API), a standard chat model representing the conventional instruction-following paradigm. This pairing enables direct comparison between reasoning-model and chat-model outputs within the same evaluation framework. The construction follows a four-stage pipeline with stringent quality controls that go beyond prior work.
	
	\textbf{Pipeline.} PubMed documents are sentence-split with NLTK~\citep{bird2009natural} and XSum on newline boundaries, filtered by a minimum sentence count, and truncated to 20 sentences with a 96-word cap per sentence. For each retained document we select $\mathrm{round}(N\gamma)$ sentences for replacement with $\gamma = 0.3$, organized into $B$ contiguous blocks with $B \sim \{1,2,3\}$ at probabilities $\{0.3, 0.5, 0.2\}$ and a one-sentence buffer between blocks. Unlike the single-span replacement used in prior work~\citep{jiang2025sendetex}, this multi-block strategy produces more varied mixing patterns that better reflect genuine human-AI co-authoring. The masked sentences are replaced with \texttt{[MASK\_$i$]} placeholders and sent to the target generator with instructions to produce one replacement per placeholder that fits naturally in context. For PubMed, an additional formatting constraint requires the generator to match the source corpus's native tokenized style: lowercase text with spaced punctuation. Prompt templates, the random seed for sentence selection, and API decoding parameters are released with the benchmark code.
	
	\textbf{Quality Filtering.} We apply three layers of filtering that are considerably stricter than existing S-AGTD benchmarks: (1)~\textit{length consistency}: each AI-generated sentence must contain at least 4 words and fall within $[0.25\times, 2.5\times]$ the average human sentence length in the same document; (2)~\textit{perplexity consistency}: the perplexity of the hybrid document, measured by Llama-3.1-8B-Instruct~\citep{grattafiori2024llama}, must not deviate from the original by more than 15 points, with both values below 150; and (3)~\textit{format integrity}: the generator output must parse into exactly the expected number of fill sentences. Samples failing any criterion are discarded and regenerated. The perplexity constraint, absent in prior benchmarks, ensures that the resulting hybrid documents are statistically coherent and cannot be trivially distinguished by simple perplexity-based detectors, forcing models to rely on deeper stylistic and structural cues.
	
	\textbf{Statistics.} Table~\ref{tab:dataset} summarizes MOSAIC. The overall mean perplexity gap between original and hybrid documents is below 0.4 points, confirming that the quality filtering produces challenging hybrid documents whose AI-generated sentences match the perplexity profile of their human-written context.
	
	\textbf{Benchmark Difficulty.} The quality controls above make MOSAIC substantially harder than prior S-AGTD benchmarks. Multi-block replacement requires detection of multiple authorship transitions per document rather than a single span, and the inclusion of reasoning-model outputs that approximate expert human writing further raises the bar.
	
	\section{SenFlow}

	Figure~\ref{fig:overview} presents an overview of SenFlow: a proxy model (\S\ref{sec:feat}) feeds parallel encoders fused via cross-attention (\S\ref{sec:repr}), followed by a GCN + CRF stage that models inter-sentence flow (\S\ref{sec:flow}). The full model is trained with a composite loss combining focal classification, InfoNCE-style contrastive, CRF negative log-likelihood, and boundary-prediction terms; the full formulation, optimizer settings, and loss weights are in Appendices~\ref{app:hyperparams} and~\ref{app:loss}.
	
	The individual building blocks of SenFlow, namely LoRA-aligned proxy features, TCN sequence encoding, graph convolution over a document graph, and CRF decoding, have antecedents in prior structured-prediction and AGTD work. SenFlow's contribution is not any single primitive but their integration under one sentence-graph formulation that treats S-AGTD as \emph{joint} label prediction over the document sentence sequence, rather than the independent per-sentence classification adopted by all prior S-AGTD methods. The ablation study (Table~\ref{tab:ablation}) bears this out: the joint inter-sentence structure, not the per-sentence encoders, is what drives the gains.
	
	\subsection{Proxy-Model Feature Extraction}
	\label{sec:feat}
	
	Following the distributional alignment strategy of DALD~\citep{zeng2024dald}, we fine-tune Llama-3.1-8B-Instruct~\citep{grattafiori2024llama} with LoRA~\citep{hu2022lora} to approximate the output distribution of the target generators, yielding a proxy model that produces more discriminative features than an unaligned base model.
	
	A key distinction from prior work is that we encode the \textit{entire document in a single forward pass} rather than processing each sentence independently. All $N$ sentences are concatenated into a single token sequence and fed through the proxy model. For each sentence $s_i$, we extract four streams by slicing the full-document outputs at token boundaries: a \textit{contextual embedding} $\mathbf{h}_i \in \mathbb{R}^{d_h}$ (mean-pooled last-layer hidden states), \textit{token probabilities} $\mathbf{p}_i \in \mathbb{R}^{L}$, \textit{token entropies} $\mathbf{e}_i \in \mathbb{R}^{L}$, and four-dimensional \textit{surprisal statistics} $\mathbf{d}_i$ (mean, variance, max, median of $-\log_2 p$). Here $d_h = 4096$ and $L = 96$.
	
	\subsection{Sentence Representation Learning}
	\label{sec:repr}
	
	\textbf{Semantic and Style Encoders.} The contextual embedding $\mathbf{h}_i$ is projected from $d_h$ to $d_{\text{model}}$ through a two-layer MLP with GELU and LayerNorm, producing a semantic representation $\mathbf{z}_i^{\text{sem}} \in \mathbb{R}^{d_{\text{model}}}$. In parallel, token probabilities and entropies are stacked into a two-channel sequence $\mathbf{S}_i \in \mathbb{R}^{L \times 2}$ and processed by a temporal convolutional network (TCN)~\citep{bai2018empirical} with exponentially dilated residual blocks; see Appendix~\ref{app:hyperparams}. The TCN captures multi-scale sequential patterns in the token-level signals, since AI-generated sentences often exhibit smoother probability trajectories than human-written ones. An attention pooling layer aggregates the temporal output into a fixed-length style vector $\mathbf{z}_i^{\text{sty}} \in \mathbb{R}^{d_{\text{model}}}$.
	
	\textbf{Dual Cross-Attention Fusion.} We compress $\mathbf{z}_i^{\text{sty}}$ and $\mathbf{z}_i^{\text{sem}}$ to $d_c$ dimensions and fuse them via bidirectional multi-head cross-attention (4 heads) with residual connections, then concatenate the projected surprisal statistics $\mathbf{d}_i$ through an MLP to produce the fused sentence representation $\mathbf{z}_i \in \mathbb{R}^{d_c}$. Full equations are given in Appendix~\ref{app:arch}. To prevent over-reliance on any single branch and improve cross-domain robustness, each compressed branch is randomly zeroed with a small probability during training, a technique we refer to as \textit{branch dropout}.
	
	\subsection{Inter-Sentence Flow Modeling}
	\label{sec:flow}
	
	\begin{figure*}[t]
		\centering
		\includegraphics[width=\textwidth]{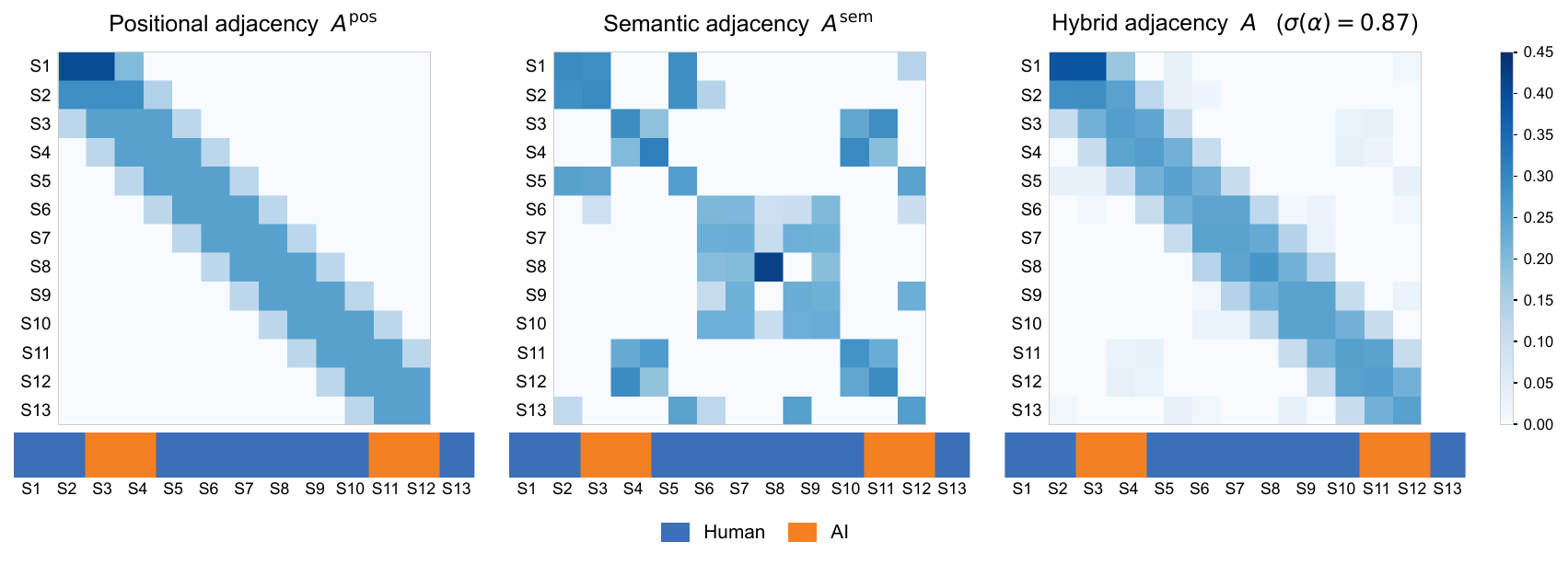}
		\caption{\textbf{The semantic branch encodes class-aligned long-range structure that local positional adjacency cannot capture.} Hybrid adjacency for a Pub$\to$XSum cross-domain test document ($N=13$); ground-truth labels below each panel, blue human, amber AI. Positional adjacency $\mathbf{A}^{\text{pos}}$ (left) is banded and local; semantic adjacency $\mathbf{A}^{\text{sem}}$ (middle) is near-perfectly class-aligned with long-range edges; the learned hybrid (right) reaches $\sigma(\alpha) = 0.87$ here, close to the Unified $0.83$. The pattern emerges on XSum, a domain absent from training.}
		\label{fig:adjacency}
	\end{figure*}
	
	\textbf{Hybrid Adjacency GCN.} Following the document-level graph paradigm developed for inter-sentence relation extraction and graph-based sentence ordering~\citep{sahu2019inter, yin2019graph}, we construct a document graph where each sentence is a node, and propagate information across nodes through a 3-layer GCN~\citep{kipf2016semi}. The design of the adjacency follows from two priors that a hybrid-document graph should encode. A \textit{locality prior} reflects that authorship transitions tend to occur in contiguous blocks rather than at isolated sentences, so adjacent sentences carry strong label correlation. A \textit{data-driven structural prior} reflects that sentences from the same author tend to cluster in representation space even when they are not adjacent in the document, providing long-range edges that local structure cannot recover. These two priors map to two complementary adjacency views, combined as a convex mixture with a single learnable weight:
	\begin{equation}
		\mathbf{A} = \sigma(\alpha) \mathbf{A}^{\text{pos}} + (1 - \sigma(\alpha)) \mathbf{A}^{\text{sem}}
	\end{equation}
	where $\alpha$ is a single learnable scalar and $\sigma$ is the sigmoid function. The \textit{positional adjacency} $\mathbf{A}^{\text{pos}}$ realizes the locality prior, with edge weights of 1.0 between immediate neighbors, 0.5 between two-hop neighbors, and self-loops included. The \textit{semantic adjacency} $\mathbf{A}^{\text{sem}}$ realizes the data-driven structural prior, constructed dynamically from the fused sentence representations: for each sentence $i$, edges are placed to its top-$k$ most similar peers in cosine space with $k=3$, retaining only positive similarities, and the resulting matrix is symmetrized. Cosine similarity is adopted as the relational metric, consistent with its standard use as a sentence-similarity measure on transformer hidden states~\citep{reimers2019sentence}. Both adjacency matrices are row-normalized before mixing. The scalar $\alpha$ is learned end-to-end together with the rest of the model (initialization in Appendix~\ref{app:hyperparams}). At convergence, $\sigma(\alpha) = 0.83$, allocating $83\%$ weight to positional adjacency and a non-trivial $17\%$ to semantic adjacency. The corresponding logit ($\approx 1.59$) lies in a high-gradient region of $\sigma$ rather than on its saturated tail, so optimization would readily drive $\sigma(\alpha)$ toward $1.0$ if the semantic branch carried no useful signal; the stably preserved $17\%$ weight indicates that the two branches contribute complementary, non-redundant information rather than the semantic branch being a spurious addition. Figure~\ref{fig:adjacency} visualizes the class-aligned long-range structure contributed by the semantic branch on a held-out cross-domain test sample. Each GCN layer applies $\mathbf{A}$ to refine node representations through a residual block of LayerNorm, GELU, and dropout (full propagation rule in Appendix~\ref{app:arch}).
	
	\textbf{CRF Sequence Decoder.} The GCN-refined representations are passed through a two-layer MLP with GELU, LayerNorm, and dropout to produce per-sentence emission scores, on top of which we apply a linear-chain conditional random field (CRF)~\citep{lafferty2001conditional}. The CRF learns transition potentials between adjacent labels, encouraging coherent label sequences and penalizing rapid oscillations between human and AI labels that rarely occur in practice. During training, the CRF contributes a negative log-likelihood loss; during inference, the Viterbi algorithm produces the globally optimal label sequence. Empirically, the learned transition matrix favors same-label continuations over cross-label transitions by a factor of roughly $2.5\times$ on both classes.
	
	\section{Experiments}
	
	\subsection{Experimental Setup}
	\label{sec:setup}
	
	\textbf{Evaluation Protocols.} We evaluate all methods under three protocols of increasing difficulty. The Unified protocol pools all four subsets PD, PK, XD, and XK for training, and each subset is evaluated independently. The Cross-generator protocol trains on one generator within a domain and tests on the other in the same domain. The Cross-domain protocol trains on one domain and tests on the other. The latter two protocols progressively probe the generalization of each method to unseen generators and unseen domains.
	
	\textbf{Baselines.} We compare SenFlow against two training-free methods, \textbf{Fast-DetectGPT}~\citep{bao2024fast} and \textbf{Binoculars}~\citep{hans2024spotting}, and three training-based S-AGTD methods, \textbf{SeqXGPT}~\citep{wang2023seqxgpt}, \textbf{POGER}~\citep{shi2024ten}, and the current state-of-the-art \textbf{SenDetEX}~\citep{jiang2025sendetex}. Training-free methods are applied at the sentence level with a single threshold tuned on the validation set; training-based methods are retrained on MOSAIC under each protocol. Detailed configurations are in Appendix~\ref{app:baselines}.
	
	\textbf{Implementation Details.} We report four metrics. Our primary metric, \textbf{Macro-F1}, is the unweighted mean of per-class F1 over human and AI. \textbf{AUC} and \textbf{MCC} complement Macro-F1 under class imbalance, and the three above are reported per subset. \textbf{Avg F1} additionally reports the mean of Macro-F1 across PD, PK, XD, and XK as a single-number summary. All performance differences are in percentage points (pp). The proxy is Llama-3.1-8B-Instruct~\citep{grattafiori2024llama} fine-tuned with LoRA~\citep{hu2022lora} under the DALD strategy~\citep{zeng2024dald}. SenFlow has approximately 1.7M trainable parameters; the full configuration, regularization stack, and loss weights are in Appendix~\ref{app:hyperparams}. Unless otherwise specified, all numbers in Tables~\ref{tab:main}, \ref{tab:cross_gen}, and~\ref{tab:cross_domain} are from runs with random seed 42.
	
	\subsection{Main Results}
	\label{sec:main_results}
	
	\begin{table*}[!t]
		\centering
		\resizebox{\textwidth}{!}{%
			\begin{tabular}{@{}ll cccc cccc cccc c@{}}
				\toprule
				& & \multicolumn{4}{c}{\textbf{Macro-F1}} & \multicolumn{4}{c}{\textbf{AUC}} & \multicolumn{4}{c}{\textbf{MCC}} & \\
				\cmidrule(lr){3-6} \cmidrule(lr){7-10} \cmidrule(lr){11-14}
				\textbf{Method} & \textbf{Type} & PD & PK & XD & XK & PD & PK & XD & XK & PD & PK & XD & XK & \textbf{Avg F1} \\
				\midrule
				Fast-DetectGPT$^\dagger$ & TF & .586 & .604 & .558 & .577 & .574 & .625 & .546 & .590 & .182 & .215 & .116 & .153 & .581 \\
				Binoculars$^\dagger$ & TF & .677 & .717 & .621 & .651 & .765 & .793 & .683 & .713 & .356 & .434 & .243 & .303 & .667 \\
				\midrule
				POGER & TB & .854 & .719 & .800 & .729 & .938 & .840 & .890 & .830 & .715 & .440 & .624 & .465 & .776 \\
				SeqXGPT & TB & .863 & .762 & .841 & .758 & .950 & .883 & .932 & .866 & .728 & .529 & .687 & .517 & .806 \\
				SenDetEX & TB & .943 & .925 & .933 & .893 & .985 & .977 & .981 & .955 & .885 & .850 & .867 & .787 & .924 \\
				\textbf{SenFlow} & TB & \textbf{.956} & \textbf{.942} & \textbf{.948} & \textbf{.915} & \textbf{.989} & \textbf{.981} & \textbf{.983} & \textbf{.968} & \textbf{.912} & \textbf{.885} & \textbf{.895} & \textbf{.831} & \textbf{.940} \\
				\bottomrule
			\end{tabular}%
		}
		\caption{\textbf{SenFlow surpasses the previous state of the art on every subset and metric under the Unified protocol.} TF = training-free, TB = training-based. Avg F1 = arithmetic mean of Macro-F1 across the four subsets. $\dagger$Training-free methods use a single threshold tuned on the combined validation set. Best results in \textbf{bold}.}
		\label{tab:main}
	\end{table*}
	
	Table~\ref{tab:main} presents the Unified protocol results. The two training-free baselines, Fast-DetectGPT and Binoculars, perform poorly at the sentence level (Avg F1 of 0.581 and 0.667), confirming that document-level detectors do not transfer to sentence-level detection. Among training-based methods, SenFlow advances the previous state of the art from SenDetEX's 0.924 to 0.940 Avg F1, leading on every subset and across all three metrics, by 1.68 pp on Avg F1.
	
	\begin{figure}[t]
		\centering
		\includegraphics[trim=0 0 60 0, clip, width=\columnwidth]{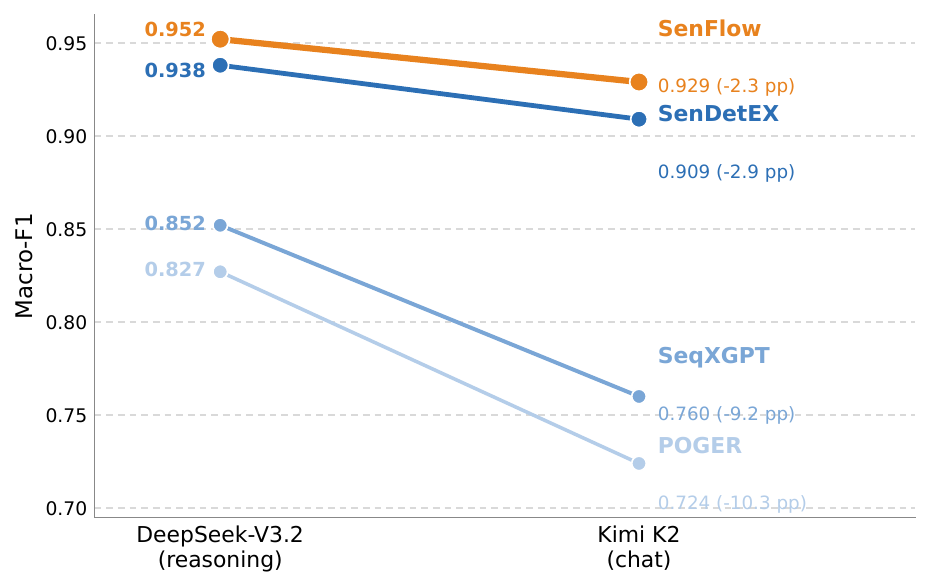}
		\caption{\textbf{Counter to intuition, DeepSeek-V3.2 insertions are easier to detect than Kimi K2 insertions, consistently across all four training-based methods.} Macro-F1 on the DeepSeek-V3.2 (reasoning) and Kimi K2 (chat) subsets under the Unified protocol, averaged across PubMed and XSum. Parenthetical values report the pp drop from DeepSeek-V3.2 to Kimi K2.}
		\label{fig:reasoning_chat}
	\end{figure}
	
	A consistent pattern across all training-based methods is that the chat-model subsets PK and XK prove harder to detect than their reasoning-model counterparts PD and XD (Figure~\ref{fig:reasoning_chat}). SenFlow exhibits the smallest drop across this shift, roughly four times smaller than POGER ($4.5\times$) and SeqXGPT ($4.0\times$). This direction is itself counterintuitive: the stronger logical coherence of the reasoning-model generator might be expected to make its insertions \emph{harder} to flag, yet on both generators evaluated the opposite holds. After MOSAIC's perplexity-consistency filter equalizes overt cues, narrowing the overall hybrid-to-original perplexity gap to under 0.4 points (Table~\ref{tab:dataset}), a \emph{generator-dependent sentence-length structural gap} remains: DeepSeek-V3.2 insertions deviate 46--52\% downward from the surrounding human prose against 32--34\% for Kimi K2 (Appendix~\ref{app:stylometric}), and this larger residual gap aligns with the easier detectability of DeepSeek-V3.2 insertions despite their surface resemblance to expert human writing. The finding is thus about the benchmark, not the paradigm: perplexity-level naturalness alone does not shield an insertion from sentence-level detection. We scope this to the two generators evaluated and discuss paradigm-level generality in the Limitations.
	
	\textbf{Significance of the Unified margin.}\label{sec:diagnostics} The gains over SenDetEX are not test-set or training-seed artifacts: a paired document-level bootstrap ($N = 2{,}000$) gives an aggregate Avg F1 delta of $+2.04$ pp $[+1.79, +2.31]$ with all five per-subset and aggregate intervals excluding zero ($p < 5 \times 10^{-4}$), and three independent seeds yield Macro-F1 $0.9413 \pm 0.0011$, a $\pm 0.1$ pp run-to-run variance far below the $1.68$ pp gap over SenDetEX.
	
	\subsection{Generalization Analysis}
	
	\begin{table*}[!t]
		\centering
		\resizebox{\textwidth}{!}{%
			\begin{tabular}{@{}l cccc cccc cccc c@{}}
				\toprule
				& \multicolumn{4}{c}{\textbf{Macro-F1}} & \multicolumn{4}{c}{\textbf{AUC}} & \multicolumn{4}{c}{\textbf{MCC}} & \\
				\cmidrule(lr){2-5} \cmidrule(lr){6-9} \cmidrule(lr){10-13}
				\textbf{Method} & PD$\to$PK & PK$\to$PD & XD$\to$XK & XK$\to$XD & PD$\to$PK & PK$\to$PD & XD$\to$XK & XK$\to$XD & PD$\to$PK & PK$\to$PD & XD$\to$XK & XK$\to$XD & \textbf{Avg F1} \\
				\midrule
				POGER    & .656 & .783 & .690 & .759 & .811 & .874 & .805 & .844 & .346 & .567 & .382 & .528 & .722 \\
				SeqXGPT  & .655 & .792 & .706 & .790 & .827 & .894 & .831 & .891 & .355 & .583 & .414 & .581 & .736 \\
				SenDetEX & .843 & .879 & .821 & .873 & .938 & .952 & .914 & .937 & .696 & .769 & .657 & .752 & .854 \\
				\textbf{SenFlow} & \textbf{.856} & \textbf{.901} & \textbf{.825} & \textbf{.889} & \textbf{.941} & \textbf{.969} & \textbf{.920} & \textbf{.963} & \textbf{.697} & \textbf{.807} & \textbf{.661} & \textbf{.784} & \textbf{.868} \\
				\bottomrule
			\end{tabular}%
		}
		\caption{\textbf{SenFlow leads on every cross-generator transfer direction.} Models are trained on one generator and tested on the other within the same domain. Avg F1 = arithmetic mean of Macro-F1 across the four transfer directions. Best results in \textbf{bold}.}
		\label{tab:cross_gen}
	\end{table*}
	
	\begin{table*}[!tp]
		\centering
		\resizebox{\textwidth}{!}{%
			\begin{tabular}{@{}l cccc cccc cccc c@{}}
				\toprule
				& \multicolumn{4}{c}{\textbf{Macro-F1}} & \multicolumn{4}{c}{\textbf{AUC}} & \multicolumn{4}{c}{\textbf{MCC}} & \\
				\cmidrule(lr){2-5} \cmidrule(lr){6-9} \cmidrule(lr){10-13}
				\textbf{Method} & Pub$\to$XD & Pub$\to$XK & XS$\to$PD & XS$\to$PK & Pub$\to$XD & Pub$\to$XK & XS$\to$PD & XS$\to$PK & Pub$\to$XD & Pub$\to$XK & XS$\to$PD & XS$\to$PK & \textbf{Avg F1} \\
				\midrule
				POGER    & .794 & .726 & .846 & .622 & .884 & .818 & .936 & .801 & .610 & .460 & .691 & .283 & .747 \\
				SeqXGPT  & .801 & .740 & .844 & .664 & .895 & .840 & .937 & .822 & .613 & .484 & .688 & .369 & .762 \\
				SenDetEX & .739 & .739 & .807 & .792 & .818 & .821 & .903 & .884 & .481 & .478 & .618 & .587 & .769 \\
				\textbf{SenFlow} & \textbf{.813} & \textbf{.759} & \textbf{.848} & \textbf{.823} & \textbf{.904} & \textbf{.856} & \textbf{.943} & \textbf{.916} & \textbf{.661} & \textbf{.522} & \textbf{.706} & \textbf{.654} & \textbf{.811} \\
				\bottomrule
			\end{tabular}%
		}
		\caption{\textbf{SenFlow's advantage widens to 4.15 pp Avg F1 under cross-domain transfer, the largest margin of the three protocols.} Models are trained on one domain and tested on the other. Pub = PubMed, XS = XSum. Best results in \textbf{bold}.}
		\label{tab:cross_domain}
	\end{table*}
	
	\textbf{Cross-Generator.} Table~\ref{tab:cross_gen} evaluates generalization to unseen generators within the same domain. SenFlow achieves an Avg F1 of 0.868, outperforming SenDetEX by 1.38 pp, with consistent leads on AUC and MCC. SenFlow further encodes the entire document in a single forward pass, while SenDetEX invokes the proxy once per sentence; with an average of 17.6 sentences per document on MOSAIC, this corresponds to approximately $20\times$ fewer proxy-model inference calls.
	
	\textbf{Cross-Domain.} Table~\ref{tab:cross_domain} presents the most challenging protocol. SenFlow achieves an Avg F1 of 0.811, outperforming SenDetEX by 4.15 pp, with consistent leads on all four transfer directions and on AUC ($+4.82$ pp) and MCC ($+9.48$ pp). The margin peaks at this hardest protocol, exceeding the 1.68 pp on Unified and 1.38 pp on Cross-generator margins, and is attained without any explicit domain adaptation. The structural prior captured by inter-sentence flow modeling appears intrinsically transferable: while surface-level stylistic signals are known to shift across domains and challenge in-the-wild detection~\citep{li2024mage}, the tendency of authorship labels to exhibit local consistency does not. That the margin is \emph{largest} precisely under cross-domain transfer also bears on a possible concern that SenFlow merely exploits the contiguous-block layout used to construct MOSAIC: a model that had overfit that layout would degrade, not improve, when tested on a domain unseen in training, whereas the inter-sentence advantage instead grows; the block-count stratification below corroborates this directly. Appendix~\ref{app:tokenization} verifies that SenFlow's margin is not confounded by surface tokenization differences between PubMed and XSum.

	\subsection{Ablation Study}
	\begin{table}[!htbp]
		\centering
		\small
		\setlength{\tabcolsep}{8pt}
		\begin{tabular}{@{}l ccc@{}}
			\toprule
			\textbf{Variant} & \textbf{Avg F1} & \textbf{Avg AUC} & \textbf{Avg MCC} \\
			\midrule
			\textbf{Full Model} & \textbf{.940} & \textbf{.980} & \textbf{.881} \\
			\quad w/o GCN       & .922 & .944 & .852 \\
			\quad w/o CRF       & .931 & .953 & .867 \\
			\quad w/o CL        & .935 & .962 & .873 \\
			\quad w/o TCN       & .932 & .958 & .871 \\
			\bottomrule
		\end{tabular}
		\caption{\textbf{The Hybrid Adjacency GCN is the dominant component across all metrics.} Ablation study under the Unified protocol, averaged across PD, PK, XD, and XK. Best results in \textbf{bold}.}
		\label{tab:ablation}
	\end{table}
	
	Table~\ref{tab:ablation} reveals a two-level structure in SenFlow's design. The Hybrid Adjacency GCN causes the largest drop on every metric, confirming inter-sentence dependencies as the primary driver. The CRF contributes the second-largest drop by enforcing label-level coherence, while the per-sentence components TCN and contrastive loss provide smaller, complementary effects, with the TCN producing a marginally larger drop than the contrastive loss (0.8 vs.\ 0.5 pp Avg F1). The GCN's dominance as the largest-drop component holds uniformly across all four subsets, with the full per-subset breakdown in Appendix~\ref{app:ablation_full}; qualitative case studies illustrating the contrast between inter-sentence flow and per-sentence classification are provided in Appendix~\ref{app:casestudy}.
	
	\textbf{Error decomposition by sentence position.} Partitioning test sentences by their ground-truth context shows that SenFlow's advantage concentrates on AI sentences embedded in human context, the hardest category, where inter-sentence propagation supplies a structural signal that per-sentence classification lacks; consistently, SenFlow's predicted per-document transition count sits closer to the ground truth than SenDetEX's. Full per-position error rates and transition counts are in Appendix~\ref{app:ablation_full}.
	
	\textbf{Robustness to insertion structure.} A natural concern is that SenFlow's locality prior and CRF decoder simply capitalize on the multi-block construction of MOSAIC. Stratifying the Unified test documents by their number of contiguous AI blocks $B$ refutes this: SenFlow leads SenDetEX in \emph{every} stratum, by $+1.62$, $+2.40$, and $+1.82$ pp Macro-F1 at $B = 1, 2, 3$ across 815, 1{,}211, and 374 documents. The lead is already substantial at $B = 1$, the single-block regime that coincides with the single-span replacement used by prior S-AGTD benchmarks; the inter-sentence advantage therefore does not arise from MOSAIC's multi-block design, but persists even where the benchmark reduces to the prior single-span setting. The lead is also non-monotonic in $B$, contrary to the steadily increasing advantage a model merely exploiting the number of blocks would produce.

	\section{Conclusion}
	
	We presented MOSAIC, a benchmark of 16{,}000 hybrid documents spanning reasoning- and chat-model generators, and SenFlow, which recasts S-AGTD as structured prediction integrating graph-based inter-sentence propagation with CRF decoding. SenFlow reaches state-of-the-art performance across three protocols, with a +4.15 pp Avg F1 margin on cross-domain transfer attained without explicit domain adaptation; the structural prior it captures transfers to both unseen generators and unseen domains. We further showed that a generator-dependent sentence-length signal survives MOSAIC's perplexity-consistency filter and remains exploitable at the sentence level, so equalizing perplexity-level naturalness does not by itself render AI insertions undetectable.
	
	\section*{Ethical Considerations}
	
	This work develops a sentence-level detector for AI-generated text, an application area with inherent dual-use considerations: the same detection capability that supports content authenticity can also enable false attribution and misuse. False positives may unfairly attribute human writing to AI sources, with potential consequences in academic and editorial settings. We therefore recommend that SenFlow be used as an auxiliary tool alongside human review rather than as a sole decision criterion, particularly in high-stakes scenarios. The benchmark released in this work is constructed from publicly available corpora, namely PubMed abstracts and BBC news summaries from XSum, without collecting or introducing personally identifying information. The AI-generated portions are produced by the listed LLMs without involving human annotators.
	
	\section*{Limitations}
	
	\textbf{Benchmark scope.} MOSAIC covers two text domains, biomedical literature and news, and two generators, DeepSeek-V3.2 and Kimi K2; all experiments are conducted on English text. Extending MOSAIC to additional domains, languages, and a broader pool of generators would further strengthen the external validity of these findings. Beyond domain and generator coverage, four further design choices constrain its scope. \textbf{(i)}~All hybrid documents are constructed at a fixed AI replacement ratio of approximately 30\%; robustness to lighter ratios in the 5--10\% range or heavier ratios in the 50--70\% range remains untested. \textbf{(ii)}~The task is binary, distinguishing human from AI sentences; extending the task to multi-class generator attribution, where each AI sentence is mapped to its source model, would enable a more fine-grained evaluation but lies outside the present scope. \textbf{(iii)}~The perplexity-consistency filter ($|\Delta\text{PPL}| \leq 15$) discards generated samples whose proxy-model perplexity diverges sharply from the surrounding human context. This is by design: it forces detectors to rely on signals beyond raw perplexity gaps. However, the retained samples may be biased toward AI outputs that approximate human-like perplexity profiles, and the relative difficulty against an unfiltered distribution should be characterized in future work. \textbf{(iv)}~The residual-signal finding (\S\ref{sec:main_results}) is established on two generators, DeepSeek-V3.2 and Kimi K2, and we scope it accordingly: the claim is that a generator-dependent length signal survives the perplexity filter and stays exploitable, not that ``reasoning models'' as a class are easier to detect. Because the two generators differ on several axes beyond reasoning-vs-chat, including post-training data, alignment recipe, and decoding defaults, attributing the gap specifically to the reasoning paradigm would require a broader generator pool, which we leave to future work.
	
	\textbf{Method scope.} SenFlow follows the DALD strategy and relies on a LoRA-aligned proxy model fine-tuned with samples from the target generators. The method is directly applicable in semi-black-box settings; extending it to fully black-box scenarios with only sparse API access remains an open challenge. The system processes documents up to 2{,}048 tokens and assumes known sentence boundaries; integrating sentence segmentation and longer-document handling into an end-to-end pipeline is a useful next step. Two additional method-scope limitations should be acknowledged. First, all baselines and SenFlow share the same LoRA-aligned proxy, which by design isolates the effect of the detection method from the choice of proxy but leaves the absolute contribution of proxy alignment itself unmeasured; the reported margins should therefore be read as method gain on top of an aligned proxy rather than total gain over an unaligned reference, and a controlled ablation with an unaligned base Llama proxy is left for future work. Second, robustness against post-hoc perturbations targeting sentence-level detectors, including paraphrase rewriting, human post-editing, grammar normalization, and sentence reordering~\citep{huang2024ai}, is not evaluated in the present scope.
	
	\textbf{Empirical scope.} The Cross-generator and Cross-domain results in Tables~\ref{tab:cross_gen} and~\ref{tab:cross_domain} use a single training seed, whereas the Unified margin is verified by both a paired document-level bootstrap ($N = 2{,}000$) and three independent seeds (\S\ref{sec:diagnostics}). The measured Unified run-to-run variance is only $\pm 0.1$ pp, one to two orders of magnitude below the $1.38$ pp Cross-generator and $4.15$ pp Cross-domain margins, so training-seed noise is an implausible explanation for the transfer results; we nonetheless report them from a single seed for compute reasons.
	
	%
	
	\section*{Acknowledgments}
	
	This work was supported by the National Science and Technology Basic Resources Survey Special Project of China under Grant 2025FY100104, the National Natural Science Foundation of China under Grant 72401263, and the Basic Research Programs of Taicang 2024 under Grant TC2024JC46.
	
	
	\bibliography{custom}

\begin{thebibliography}{41}
\providecommand{\natexlab}[1]{#1}

\bibitem[{Bai et~al.(2018)Bai, Kolter, and Koltun}]{bai2018empirical}
Shaojie Bai, J~Zico Kolter, and Vladlen Koltun. 2018.
\newblock An empirical evaluation of generic convolutional and recurrent
  networks for sequence modeling.
\newblock \emph{arXiv preprint arXiv:1803.01271}.

\bibitem[{Bao et~al.(2024)Bao, Zhao, Teng, Yang, and Zhang}]{bao2024fast}
Guangsheng Bao, Yanbin Zhao, Zhiyang Teng, Linyi Yang, and Yue Zhang. 2024.
\newblock Fast-{D}etect{GPT}: Efficient zero-shot detection of
  machine-generated text via conditional probability curvature.
\newblock In \emph{International Conference on Learning Representations}.

\bibitem[{Bird et~al.(2009)Bird, Klein, and Loper}]{bird2009natural}
Steven Bird, Ewan Klein, and Edward Loper. 2009.
\newblock \emph{Natural language processing with {P}ython: analyzing text with
  the natural language toolkit}.
\newblock O'Reilly Media, Inc.

\bibitem[{Cohan et~al.(2018)Cohan, Dernoncourt, Kim, Bui, Kim, Chang, and
  Goharian}]{cohan2018discourse}
Arman Cohan, Franck Dernoncourt, Doo~Soon Kim, Trung Bui, Seokhwan Kim, Walter
  Chang, and Nazli Goharian. 2018.
\newblock A discourse-aware attention model for abstractive summarization of
  long documents.
\newblock In \emph{Proceedings of the 2018 Conference of the North American
  Chapter of the Association for Computational Linguistics: Human Language
  Technologies}, volume~2, pages 615--621.

\bibitem[{Grattafiori et~al.(2024)Grattafiori, Dubey, Jauhri, Pandey, Kadian,
  Al-Dahle, Letman, Mathur, Schelten, Vaughan et~al.}]{grattafiori2024llama}
Aaron Grattafiori, Abhimanyu Dubey, Abhinav Jauhri, Abhinav Pandey, Abhishek
  Kadian, Ahmad Al-Dahle, Aiesha Letman, Akhil Mathur, Alan Schelten, Alex
  Vaughan, and 1 others. 2024.
\newblock The {L}lama 3 herd of models.
\newblock \emph{arXiv preprint arXiv:2407.21783}.

\bibitem[{Guo et~al.(2025)Guo, Yang, Zhang, Song, Wang, Zhu, Xu, Zhang, Ma, Bi
  et~al.}]{guo2025deepseek}
Daya Guo, Dejian Yang, Haowei Zhang, Junxiao Song, Peiyi Wang, Qihao Zhu,
  Runxin Xu, Ruoyu Zhang, Shirong Ma, Xiao Bi, and 1 others. 2025.
\newblock {D}eep{S}eek-{R}1: Incentivizing reasoning capability in {LLM}s via
  reinforcement learning.
\newblock \emph{arXiv preprint arXiv:2501.12948}.

\bibitem[{Hans et~al.(2024)Hans, Schwarzschild, Cherepanova, Kazemi, Saha,
  Goldblum, Geiping, and Goldstein}]{hans2024spotting}
Abhimanyu Hans, Avi Schwarzschild, Valeriia Cherepanova, Hamid Kazemi,
  Aniruddha Saha, Micah Goldblum, Jonas Geiping, and Tom Goldstein. 2024.
\newblock Spotting {LLM}s with binoculars: Zero-shot detection of
  machine-generated text.
\newblock In \emph{Proceedings of the 41st International Conference on Machine
  Learning}, pages 17519--17537.

\bibitem[{Hu et~al.(2022)Hu, Shen, Wallis, Allen-Zhu, Li, Wang, Wang, Chen
  et~al.}]{hu2022lora}
Edward~J Hu, Yelong Shen, Phillip Wallis, Zeyuan Allen-Zhu, Yuanzhi Li, Shean
  Wang, Liang Wang, Weizhu Chen, and 1 others. 2022.
\newblock Lo{RA}: Low-rank adaptation of large language models.
\newblock In \emph{International Conference on Learning Representations}.

\bibitem[{Hu et~al.(2023)Hu, Chen, and Ho}]{hu2023radar}
Xiaomeng Hu, Pin-Yu Chen, and Tsung-Yi Ho. 2023.
\newblock {RADAR}: Robust {AI}-text detection via adversarial learning.
\newblock \emph{Advances in Neural Information Processing Systems},
  36:15077--15095.

\bibitem[{Huang et~al.(2024)Huang, Zhang, Li, You, Wang, and
  Yang}]{huang2024ai}
Guanhua Huang, Yuchen Zhang, Zhe Li, Yongjian You, Mingze Wang, and Zhouwang
  Yang. 2024.
\newblock Are {AI}-generated text detectors robust to adversarial
  perturbations?
\newblock In \emph{Proceedings of the 62nd Annual Meeting of the Association
  for Computational Linguistics}, volume~1, pages 6005--6024.

\bibitem[{Jaech et~al.(2024)Jaech, Kalai, Lerer, Richardson, El-Kishky, Low,
  Helyar, Madry, Beutel, Carney et~al.}]{jaech2024openai}
Aaron Jaech, Adam Kalai, Adam Lerer, Adam Richardson, Ahmed El-Kishky, Aiden
  Low, Alec Helyar, Aleksander Madry, Alex Beutel, Alex Carney, and 1 others.
  2024.
\newblock {O}pen{AI} o1 system card.
\newblock \emph{arXiv preprint arXiv:2412.16720}.

\bibitem[{Jiang et~al.(2025)Jiang, Wu, and Zheng}]{jiang2025sendetex}
Lei Jiang, Desheng Wu, and Xiaolong Zheng. 2025.
\newblock {SenDetEX}: Sentence-level {AI}-generated text detection for
  human-{AI} hybrid content via style and context fusion.
\newblock In \emph{Proceedings of the 2025 Conference on Empirical Methods in
  Natural Language Processing}, pages 5287--5302.

\bibitem[{Kipf and Welling(2016)}]{kipf2016semi}
Thomas~N Kipf and Max Welling. 2016.
\newblock Semi-supervised classification with graph convolutional networks.
\newblock \emph{arXiv preprint arXiv:1609.02907}.

\bibitem[{Kirchenbauer et~al.(2023)Kirchenbauer, Geiping, Wen, Katz, Miers, and
  Goldstein}]{kirchenbauer2023watermark}
John Kirchenbauer, Jonas Geiping, Yuxin Wen, Jonathan Katz, Ian Miers, and Tom
  Goldstein. 2023.
\newblock A watermark for large language models.
\newblock In \emph{International Conference on Machine Learning}, pages
  17061--17084. PMLR.

\bibitem[{Lafferty et~al.(2001)Lafferty, McCallum, and
  Pereira}]{lafferty2001conditional}
John Lafferty, Andrew McCallum, and Fernando~CN Pereira. 2001.
\newblock Conditional random fields: Probabilistic models for segmenting and
  labeling sequence data.
\newblock In \emph{Proceedings of the Eighteenth International Conference on
  Machine Learning}, pages 282--289.

\bibitem[{Lee et~al.(2022)Lee, Liang, and Yang}]{lee2022coauthor}
Mina Lee, Percy Liang, and Qian Yang. 2022.
\newblock {C}o{A}uthor: Designing a human-{AI} collaborative writing dataset
  for exploring language model capabilities.
\newblock In \emph{Proceedings of the 2022 CHI Conference on Human Factors in
  Computing Systems}, pages 1--19.

\bibitem[{Li et~al.(2024)Li, Li, Cui, Bi, Wang, Wang, Yang, Shi, and
  Zhang}]{li2024mage}
Yafu Li, Qintong Li, Leyang Cui, Wei Bi, Zhilin Wang, Longyue Wang, Linyi Yang,
  Shuming Shi, and Yue Zhang. 2024.
\newblock {MAGE}: Machine-generated text detection in the wild.
\newblock In \emph{Proceedings of the 62nd Annual Meeting of the Association
  for Computational Linguistics}, volume~1, pages 36--53.

\bibitem[{Lin et~al.(2017)Lin, Goyal, Girshick, He, and
  Doll{\'a}r}]{lin2017focal}
Tsung-Yi Lin, Priya Goyal, Ross Girshick, Kaiming He, and Piotr Doll{\'a}r.
  2017.
\newblock Focal loss for dense object detection.
\newblock In \emph{Proceedings of the IEEE International Conference on Computer
  Vision}, pages 2980--2988.

\bibitem[{Loshchilov and Hutter(2016)}]{loshchilov2016sgdr}
Ilya Loshchilov and Frank Hutter. 2016.
\newblock {SGDR}: Stochastic gradient descent with warm restarts.
\newblock \emph{arXiv preprint arXiv:1608.03983}.

\bibitem[{Loshchilov and Hutter(2017)}]{loshchilov2017decoupled}
Ilya Loshchilov and Frank Hutter. 2017.
\newblock Decoupled weight decay regularization.
\newblock \emph{arXiv preprint arXiv:1711.05101}.

\bibitem[{Mao et~al.(2024)Mao, Vondrick, Wang, and Yang}]{mao2024raidar}
Chengzhi Mao, Carl Vondrick, Hao Wang, and Junfeng Yang. 2024.
\newblock {RAIDAR}: generative {AI} detection via rewriting.
\newblock \emph{arXiv preprint arXiv:2401.12970}.

\bibitem[{Mitchell et~al.(2023)Mitchell, Lee, Khazatsky, Manning, and
  Finn}]{mitchell2023detectgpt}
Eric Mitchell, Yoonho Lee, Alexander Khazatsky, Christopher~D Manning, and
  Chelsea Finn. 2023.
\newblock {D}etect{GPT}: Zero-shot machine-generated text detection using
  probability curvature.
\newblock In \emph{International Conference on Machine Learning}, pages
  24950--24962. PMLR.

\bibitem[{Narayan et~al.(2018)Narayan, Cohen, and Lapata}]{narayan2018don}
Shashi Narayan, Shay~B Cohen, and Mirella Lapata. 2018.
\newblock Don't give me the details, just the summary! topic-aware
  convolutional neural networks for extreme summarization.
\newblock In \emph{Proceedings of the 2018 Conference on Empirical Methods in
  Natural Language Processing}, pages 1797--1807.

\bibitem[{Nguyen-Son et~al.(2024)Nguyen-Son, Dao, and
  Zettsu}]{nguyen2024simllm}
Hoang-Quoc Nguyen-Son, Minh-Son Dao, and Koji Zettsu. 2024.
\newblock {S}im{LLM}: Detecting sentences generated by large language models
  using similarity between the generation and its re-generation.
\newblock In \emph{Proceedings of the 2024 Conference on Empirical Methods in
  Natural Language Processing}, pages 22340--22352.

\bibitem[{Oord et~al.(2018)Oord, Li, and Vinyals}]{oord2018representation}
Aaron van~den Oord, Yazhe Li, and Oriol Vinyals. 2018.
\newblock Representation learning with contrastive predictive coding.
\newblock \emph{arXiv preprint arXiv:1807.03748}.

\bibitem[{Reimers and Gurevych(2019)}]{reimers2019sentence}
Nils Reimers and Iryna Gurevych. 2019.
\newblock Sentence-{BERT}: Sentence embeddings using siamese {BERT}-networks.
\newblock In \emph{Proceedings of the 2019 Conference on Empirical Methods in
  Natural Language Processing and the 9th International Joint Conference on
  Natural Language Processing}, pages 3982--3992.

\bibitem[{Sadasivan et~al.(2023)Sadasivan, Kumar, Balasubramanian, Wang, and
  Feizi}]{sadasivan2023can}
Vinu~Sankar Sadasivan, Aounon Kumar, Sriram Balasubramanian, Wenxiao Wang, and
  Soheil Feizi. 2023.
\newblock Can {AI}-generated text be reliably detected?
\newblock \emph{arXiv preprint arXiv:2303.11156}.

\bibitem[{Sahu et~al.(2019)Sahu, Christopoulou, Miwa, and
  Ananiadou}]{sahu2019inter}
Sunil~Kumar Sahu, Fenia Christopoulou, Makoto Miwa, and Sophia Ananiadou. 2019.
\newblock Inter-sentence relation extraction with document-level graph
  convolutional neural network.
\newblock In \emph{Proceedings of the 57th Annual Meeting of the Association
  for Computational Linguistics}, pages 4309--4316.

\bibitem[{Shi et~al.(2024)Shi, Sheng, Cao, Mi, Hu, and Wang}]{shi2024ten}
Yuhui Shi, Qiang Sheng, Juan Cao, Hao Mi, Beizhe Hu, and Danding Wang. 2024.
\newblock Ten words only still help: Improving black-box {AI}-generated text
  detection via proxy-guided efficient re-sampling.
\newblock In \emph{Proceedings of the Thirty-Third International Joint
  Conference on Artificial Intelligence}, pages 494--502.

\bibitem[{Solaiman et~al.(2019)Solaiman, Brundage, Clark, Askell, Herbert-Voss,
  Wu, Radford, Krueger, Kim, Kreps et~al.}]{solaiman2019release}
Irene Solaiman, Miles Brundage, Jack Clark, Amanda Askell, Ariel Herbert-Voss,
  Jeff Wu, Alec Radford, Gretchen Krueger, Jong~Wook Kim, Sarah Kreps, and 1
  others. 2019.
\newblock Release strategies and the social impacts of language models.
\newblock \emph{arXiv preprint arXiv:1908.09203}.

\bibitem[{Su et~al.(2025)Su, Wang, Wan, Zhang, and Luo}]{su2025haco}
Zhixiong Su, Yichen Wang, Herun Wan, Zhaohan Zhang, and Minnan Luo. 2025.
\newblock {HACo}-{D}et: A study towards fine-grained machine-generated text
  detection under human-{AI} coauthoring.
\newblock In \emph{Proceedings of the 63rd Annual Meeting of the Association
  for Computational Linguistics}, volume~1, pages 22015--22036.

\bibitem[{Szegedy et~al.(2016)Szegedy, Vanhoucke, Ioffe, Shlens, and
  Wojna}]{szegedy2016rethinking}
Christian Szegedy, Vincent Vanhoucke, Sergey Ioffe, Jon Shlens, and Zbigniew
  Wojna. 2016.
\newblock Rethinking the inception architecture for computer vision.
\newblock In \emph{Proceedings of the IEEE Conference on Computer Vision and
  Pattern Recognition}, pages 2818--2826.

\bibitem[{Wang et~al.(2023)Wang, Li, Ren, Jiang, Zhang, and
  Qiu}]{wang2023seqxgpt}
Pengyu Wang, Linyang Li, Ke~Ren, Botian Jiang, Dong Zhang, and Xipeng Qiu.
  2023.
\newblock {SeqXGPT}: Sentence-level {AI}-generated text detection.
\newblock In \emph{Proceedings of the 2023 Conference on Empirical Methods in
  Natural Language Processing}, pages 1144--1156.

\bibitem[{Weber-Wulff et~al.(2023)Weber-Wulff, Anohina-Naumeca, Bjelobaba,
  Folt{\`y}nek, Guerrero-Dib, Popoola, {\v{S}}igut, and
  Waddington}]{weber2023testing}
Debora Weber-Wulff, Alla Anohina-Naumeca, Sonja Bjelobaba, Tom{\'a}{\v{s}}
  Folt{\`y}nek, Jean Guerrero-Dib, Olumide Popoola, Petr {\v{S}}igut, and Lorna
  Waddington. 2023.
\newblock Testing of detection tools for {AI}-generated text.
\newblock \emph{International Journal for Educational Integrity}, 19(1):1--39.

\bibitem[{Wu et~al.(2025)Wu, Yang, Zhan, Yuan, Chao, and Wong}]{wu2025survey}
Junchao Wu, Shu Yang, Runzhe Zhan, Yulin Yuan, Lidia~Sam Chao, and Derek~Fai
  Wong. 2025.
\newblock A survey on {LLM}-generated text detection: Necessity, methods, and
  future directions.
\newblock \emph{Computational Linguistics}, 51(1):275--338.

\bibitem[{Yang et~al.(2024)Yang, Cheng, Wu, Petzold, Wang, and
  Chen}]{yang2024dna}
Xianjun Yang, Wei Cheng, Yue Wu, Linda Petzold, William Wang, and Haifeng Chen.
  2024.
\newblock {DNA}-{GPT}: Divergent n-gram analysis for training-free detection of
  {GPT}-generated text.
\newblock In \emph{International Conference on Learning Representations}.

\bibitem[{Yin et~al.(2019)Yin, Song, Su, Zeng, Zhou, and Luo}]{yin2019graph}
Yongjing Yin, Linfeng Song, Jinsong Su, Jiali Zeng, Chulun Zhou, and Jiebo Luo.
  2019.
\newblock Graph-based neural sentence ordering.
\newblock \emph{arXiv preprint arXiv:1912.07225}.

\bibitem[{Zeng et~al.(2024)Zeng, Tang, Yang, Chen, Sun, Xu, Li, Chen, Cheng,
  and Xu}]{zeng2024dald}
Cong Zeng, Shengkun Tang, Xianjun Yang, Yuanzhou Chen, Yiyou Sun, Zhiqiang Xu,
  Yao Li, Haifeng Chen, Wei Cheng, and Dongkuan Xu. 2024.
\newblock {DALD}: Improving logits-based detector without logits from black-box
  {LLM}s.
\newblock \emph{Advances in Neural Information Processing Systems},
  37:54947--54973.

\bibitem[{Zhang et~al.(2024)Zhang, Gao, Chen, Huang, Huang, Sun, Zhang, Li, Fu,
  Wan et~al.}]{zhang2024llm}
Qihui Zhang, Chujie Gao, Dongping Chen, Yue Huang, Yixin Huang, Zhenyang Sun,
  Shilin Zhang, Weiye Li, Zhengyan Fu, Yao Wan, and 1 others. 2024.
\newblock {LLM}-as-a-coauthor: Can mixed human-written and machine-generated
  text be detected?
\newblock In \emph{Findings of the Association for Computational Linguistics:
  NAACL 2024}, pages 409--436.

\bibitem[{Zhao et~al.(2023{\natexlab{a}})Zhao, Zhou, Li, Tang, Wang, Hou, Min,
  Zhang, Zhang, Dong et~al.}]{zhao2023survey}
Wayne~Xin Zhao, Kun Zhou, Junyi Li, Tianyi Tang, Xiaolei Wang, Yupeng Hou,
  Yingqian Min, Beichen Zhang, Junjie Zhang, Zican Dong, and 1 others.
  2023{\natexlab{a}}.
\newblock A survey of large language models.
\newblock \emph{arXiv preprint arXiv:2303.18223}.

\bibitem[{Zhao et~al.(2023{\natexlab{b}})Zhao, Ananth, Li, and
  Wang}]{zhao2023provable}
Xuandong Zhao, Prabhanjan Ananth, Lei Li, and Yu-Xiang Wang.
  2023{\natexlab{b}}.
\newblock Provable robust watermarking for {AI}-generated text.
\newblock \emph{arXiv preprint arXiv:2306.17439}.

\end{thebibliography}

	\appendix 
	
	\begin{table*}[!t]
		\centering
		\small
		\setlength{\tabcolsep}{12pt}
		\begin{tabular}{@{}lll@{}}
			\toprule
			\textbf{Group} & \textbf{Component} & \textbf{Value} \\
			\midrule
			\multirow{7}{*}{\textit{Architecture}}
			& $d_{\text{model}}$              & 128 \\
			& $d_c$ (compressed dimension)    & 96 \\
			& TCN levels                      & 4 \\
			& TCN kernel size                 & 3 \\
			& TCN dilation rates              & 1, 2, 4, 8 \\
			& GCN layers                      & 3 \\
			& Hybrid adjacency $\sigma(\alpha)$ initialization & 0.88 \\
			\midrule
			\multirow{2}{*}{\textit{Inter-Sentence Flow}}
			& Semantic adjacency top-$k$      & 3 \\
			& CRF transition initialization   & uniform \\
			\midrule
			\multirow{8}{*}{\textit{Optimization}}
			& Optimizer                       & AdamW \\
			& Learning rate                   & $3 \times 10^{-4}$ \\
			& Weight decay                    & 0.01 \\
			& Batch size                      & 48 \\
			& Max epochs                      & 50 \\
			& Early stopping patience         & 10 \\
			& LR schedule                     & 3-epoch warmup followed by cosine decay \\
			& Gradient clipping (max norm)    & 2.0 \\
			\midrule
			\multirow{6}{*}{\textit{Loss Weights}}
			& Contrastive loss $\alpha_{\text{cl}}$           & 0.05 \\
			& CRF loss $\alpha_{\text{crf}}$                  & 0.5 \\
			& Boundary-prediction loss $\alpha_{\text{pos}}$  & 0.1 \\
			& Focal loss $\gamma$ / $\alpha$                  & 2.0 / 0.5 \\
			& Label smoothing                                 & 0.10 \\
			& Contrastive loss warmup                         & 2 epochs \\
			\midrule
			\multirow{8}{*}{\textit{Regularization}}
			& Branch dropout                  & 0.15 \\
			& GCN dropout                     & 0.15 \\
			& Classification head dropout     & 0.20 \\
			& EMA decay                       & 0.999 \\
			& Manifold mixup $\alpha$         & 0.2 \\
			& Manifold mixup probability      & 0.5 \\
			& AWP learning rate / $\epsilon$  & $1 \times 10^{-3}$ / $1 \times 10^{-3}$ \\
			& AWP start epoch                 & 5 \\
			\bottomrule
		\end{tabular}
		\caption{Complete hyperparameter settings for SenFlow. The same configuration is used across all three evaluation protocols without protocol-specific tuning. AWP = adversarial weight perturbation; EMA = exponential moving average over model weights.}
		\label{tab:hyperparams}
	\end{table*}
	
	\section{Network Architecture Equations}
	\label{app:arch}
	
	This appendix provides the detailed equations for the Dual Cross-Attention Fusion (\S\ref{sec:repr}) and the Hybrid Adjacency GCN propagation (\S\ref{sec:flow}).
	
	\textbf{Dual Cross-Attention Fusion.} The compressed style and semantic representations $\mathbf{z}^{\text{sty}}_c, \mathbf{z}^{\text{sem}}_c \in \mathbb{R}^{d_c}$ are fused via bidirectional multi-head cross-attention with residual connections:
	\begin{equation}
		\begin{aligned}
			\mathbf{a}_1 &= \text{MHA}(\mathbf{z}^{\text{sty}}_c, \mathbf{z}^{\text{sem}}_c, \mathbf{z}^{\text{sem}}_c), \\
			\mathbf{a}_2 &= \text{MHA}(\mathbf{z}^{\text{sem}}_c, \mathbf{z}^{\text{sty}}_c, \mathbf{z}^{\text{sty}}_c)
		\end{aligned}
	\end{equation}
	\begin{equation}
		\mathbf{z}^{\text{cross}}_i = \text{LN}([\mathbf{a}_1; \mathbf{a}_2]\mathbf{W}_f + \mathbf{z}^{\text{sty}}_c)
	\end{equation}
	where MHA denotes multi-head attention with 4 heads and LN is LayerNorm. The surprisal statistics $\mathbf{d}_i$ are then projected to $d_c$ dimensions and concatenated with $\mathbf{z}^{\text{cross}}_i$ before a final MLP fusion:
	\begin{equation}
		\mathbf{z}_i = \text{MLP}([\mathbf{z}^{\text{cross}}_i; \mathbf{d}_i\mathbf{W}_d]) \in \mathbb{R}^{d_c}
	\end{equation}
	
	\textbf{GCN Propagation.} Each layer of the Hybrid Adjacency GCN follows a residual propagation rule:
	\begin{equation}
		\mathbf{Z}^{(l+1)} = \mathbf{Z}^{(l)} + \phi\!\left(\mathbf{A} \mathbf{Z}^{(l)} \mathbf{W}^{(l)}\right)
	\end{equation}
	where $\phi(\cdot)$ denotes the composition of LayerNorm, GELU activation, and dropout, and $\mathbf{A}$ is the hybrid adjacency defined in \S\ref{sec:flow}.
	
	\section{SenFlow Implementation Details}
	\label{app:hyperparams}
	
	Table~\ref{tab:hyperparams} lists the complete set of hyperparameters used to train SenFlow under all three protocols. The same configuration is used across the Unified, Cross-generator, and Cross-domain settings, with no protocol-specific tuning. All experiments are conducted on a single NVIDIA RTX PRO 6000 GPU with 96 GB memory.

	\section{Training Objective Details}
	\label{app:loss}
	
	The total training loss combines four terms:
	\begin{equation}
		\mathcal{L} = \mathcal{L}_{\text{focal}} + \alpha_{\text{cl}} \mathcal{L}_{\text{cl}} + \alpha_{\text{crf}} \mathcal{L}_{\text{crf}} + \alpha_{\text{pos}} \mathcal{L}_{\text{pos}}
	\end{equation}
	
	\textbf{Focal Loss.} The primary classification loss uses focal weighting~\citep{lin2017focal} with label smoothing~\citep{szegedy2016rethinking} to address class imbalance between human and AI sentences.
	
	\textbf{Style Contrastive Loss.} An InfoNCE-based contrastive loss~\citep{oord2018representation} is applied to the style representations $\mathbf{z}^{\text{sty}}$, encouraging sentences with the same label to cluster together in the style space while pushing apart sentences with different labels. This loss is activated after a 2-epoch warmup, while the others are active throughout training.
	
	\textbf{CRF Loss.} The negative log-likelihood of the ground-truth label sequence under the linear-chain CRF, encouraging globally consistent predictions.
	
	\textbf{Boundary-Prediction Loss.} An auxiliary classification head attached to the GCN output predicts, for each sentence, whether it is a document boundary. Each sentence is assigned one of three classes based on its position in the document: \textit{begin} for the first sentence, \textit{end} for the last, and \textit{middle} for all others. This auxiliary supervision encourages the GCN to retain sensitivity to document-level position, providing a complementary structural signal alongside the main classification objective. The loss is a standard cross-entropy term.
	
	\textbf{Optimizer.} SenFlow is trained with AdamW~\citep{loshchilov2017decoupled} using linear warmup followed by cosine decay~\citep{loshchilov2016sgdr}. Detailed values are listed in Table~\ref{tab:hyperparams}.
	
	\section{Baseline Configurations}
	\label{app:baselines}
	
	For fair comparison, all baselines use the same LoRA fine-tuned Llama-3.1-8B-Instruct as their proxy or scoring model, identical to the proxy model used by SenFlow. This isolates the effect of the detection method itself from the choice of underlying language model. The specific configurations are described below.
	
	\textbf{Fast-DetectGPT.} We follow the original sampling-based scoring procedure, using our LoRA-aligned Llama-3.1-8B-Instruct as the scoring model. Each sentence is scored independently and assigned a label based on a single threshold. The threshold is selected by grid search on the validation set to maximize Macro-F1, with higher scores corresponding to AI-generated text.
	
	\textbf{Binoculars.} The contrasting score is computed using two related models: the scoring model is our LoRA-aligned Llama-3.1-8B-Instruct, and the reference model is the original Llama-3.1-8B-Instruct without LoRA adaptation. The threshold is tuned on the validation set in the same manner as Fast-DetectGPT, with lower scores corresponding to AI-generated text.
	
	\textbf{SeqXGPT.} We reproduce SeqXGPT~\citep{wang2023seqxgpt} as a word-level sequence labeling model over token-level log probability and entropy features extracted from our proxy model. The architecture consists of a 2-channel input projection, two convolutional layers with kernel sizes 5 and 3, and a 3-layer Transformer encoder with 4 attention heads and feed-forward dimension 512, with $d_{\text{model}} = 256$ and dropout 0.1. Per-token predictions are aggregated to sentence labels via majority vote. The model is trained for up to 30 epochs with AdamW at learning rate $2 \times 10^{-4}$ and weight decay 0.01, batch size 256, and early stopping with patience 8.
	
	\textbf{POGER.} We reproduce POGER~\citep{shi2024ten} by selecting the top-$K$ ($K=16$) representative tokens per sentence ranked by surprisal, extracting their probability, entropy, and surprisal values, and concatenating them with the four-dimensional surprisal statistics. A 3-layer MLP classifier with $d_{\text{model}} = 128$ produces the final logits. The original POGER paper recommends $K=10$ for document-level detection; we re-tuned $K$ on the MOSAIC validation set across $\{10, 16, 24\}$ at sentence granularity, and $K=16$ yielded the strongest validation Macro-F1, so we adopt it throughout to avoid weakening the POGER baseline. Training follows the same schedule as SeqXGPT but with learning rate $3 \times 10^{-4}$ and batch size 512.
	
	\textbf{SenDetEX.} We reproduce SenDetEX~\citep{jiang2025sendetex} including a style extractor based on a 2-layer Transformer over convolutional features with $d_{\text{model}} = 256$, a triple cross-attention fusion module, and a classification head. The intrinsic and inferred semantic embeddings are obtained from the 4096-dimensional final hidden states of our proxy model. Training uses AdamW with learning rate $2 \times 10^{-4}$, batch size 256, weight decay 0.01, and early stopping with patience 8 over up to 30 epochs.
	
	\section{Stylometric Analysis of Reasoning vs.\ Chat Model Outputs}
	\label{app:stylometric}
	\begin{table}[!t]
		\centering
		\small
		\setlength{\tabcolsep}{10pt}
		\begin{tabular}{@{}l ccc@{}}
			\toprule
			\textbf{Subset} & \textbf{Human mean} & \textbf{AI mean} & \textbf{Downward dev.} \\
			\midrule
			PD & 27.09 & 12.98 & 52.1\% \\
			PK & 27.63 & 18.22 & 34.0\% \\
			XD & 23.37 & 12.53 & 46.4\% \\
			XK & 23.61 & 15.96 & 32.4\% \\
			\bottomrule
		\end{tabular}
		\caption{Average sentence length in words for Human and AI sentences across the four MOSAIC subsets defined in Table~\ref{tab:dataset}. Downward deviation $= |\text{AI}_{\text{mean}} - \text{Hum}_{\text{mean}}| / \text{Hum}_{\text{mean}}$. DeepSeek (reasoning) generates substantially shorter sentences than Kimi (chat), yielding a larger structural gap against the surrounding human prose in both source domains.}
		\label{tab:stylometric}
	\end{table}
	
	\begin{table*}[h]
		\centering
		\footnotesize
		\setlength{\tabcolsep}{6pt}
		\renewcommand{\arraystretch}{1.2}
		\begin{tabular}{@{}c p{0.75\textwidth} c c c@{}}
			\toprule
			\textbf{\#} & \textbf{Sentence} & \textbf{GT} & \textbf{SF} & \textbf{SD} \\
			\midrule
			1 & Murray Geddes, 37, from Aberlour in Moray, admitted causing the death of passenger Graeme McKenzie, also 37, by driving dangerously on the A941 Craigellachie to Rothes road in May. & H & H & H \\
			2 & Mr McKenzie, an offshore worker from Rothes, was thrown from the vehicle. & H & H & A$^{\times}$ \\
			3 & Sentence at the High Court in Edinburgh was deferred. & H & H & H \\
			4 & The court heard offshore driller Geddes had a speeding conviction from March for driving at 93mph in a 60mph zone on the A96 near Huntly. & H & H & H \\
			5 & During the hearing, the court heard details of the fatal collision. & \textbf{A} & A & A \\
			6 & Advocate depute Andrew Brown QC said: ``The accused and the deceased were close friends who grew up together in the Speyside area.'' & H & H & A$^{\times}$ \\
			7 & A motorist who was overtaken by Geddes later said: ``I would say it was like a speed demon.'' & H & H & H \\
			8 & The vehicle collided with a roadside & \textbf{A} & A & A \\
			9 & Geddes' previous conviction was taken into account during the proceedings. & \textbf{A} & A & H$^{\times}$ \\
			10 & The judge deferred sentence on Geddes for the preparation of a background report and he was remanded in custody. & H & H & H \\
			\bottomrule
		\end{tabular}
		\caption{Case A: a hybrid news document from XSum + DeepSeek. GT denotes the ground-truth label, with H for Human and \textbf{A} for AI. SF and SD denote the SenFlow and SenDetEX predictions; superscript $\times$ marks an incorrect prediction. SenFlow classifies all ten sentences correctly. SenDetEX produces two false positives at sentences 2 and 6 and one false negative at sentence 9.}
		\label{tab:case_a}
	\end{table*}
	
	To characterize the residual signal that survives MOSAIC's perplexity-consistency filter (\S\ref{sec:main_results}), we measure the sentence-length distribution of AI-generated sentences relative to the surrounding human prose in each MOSAIC subset. For each document the word count is averaged across AI sentences and across Human sentences separately, then averaged over documents within each subset. Table~\ref{tab:stylometric} reports both means and the absolute downward deviation of AI from Human, defined as $|\text{AI}_{\text{mean}} - \text{Hum}_{\text{mean}}| / \text{Hum}_{\text{mean}}$.
	
	DeepSeek sentences are systematically shorter than the surrounding human prose, with downward deviations of 52.1\% on PubMed and 46.4\% on XSum. Kimi sentences exhibit a markedly smaller deviation of 34.0\% and 32.4\% in the same two domains. The pattern is consistent across both source domains and is compatible with reasoning-model insertions forming a structurally distinct block within the hybrid document, providing a candidate sentence-level cue that detectors may exploit even when surface-level vocabulary is otherwise close to human prose. The analysis is correlational rather than causal: it does not isolate the contribution of the length gap from other stylistic factors that differ between the two generators, and a controlled comparison that matches sentence-length distributions across DeepSeek and Kimi outputs would be required to establish a causal link.
	
	Two within-AI lexical metrics were also computed for completeness, namely type-token ratio and mean pairwise Jaccard similarity over AI-sentence token sets. DeepSeek shows higher token-type richness and lower pairwise Jaccard than Kimi, in line with the terminological precision typical of reasoning-model outputs. Among the metrics computed here, the length-distribution structural gap is the one most strongly aligned with the observed detectability pattern, while within-AI vocabulary metrics show no comparable separation.

	\section{Cross-Domain Margin and Surface Tokenization}
	\label{app:tokenization}
	
	PubMed and XSum differ in surface tokenization, and this difference is a property of the source corpora rather than a preprocessing choice introduced during construction: the PubMed corpus released by \citet{cohan2018discourse} is distributed in pre-tokenized form, namely lowercase text with spaced punctuation, while XSum~\citep{narayan2018don} retains standard mixed-case prose. The AI-generated portions are required to match each domain's native style so that surface form cannot serve as a cue for authorship within a domain (\S\ref{sec:bench}). In cross-domain transfer the model therefore sees a tokenization shift in addition to the domain shift, raising the question of whether SenFlow's $4.15$ pp Avg F1 advantage over SenDetEX (Table~\ref{tab:cross_domain}) reflects robustness to that surface shift rather than to genuine domain shift.
	
	\begin{table*}[h]
		\centering
		\footnotesize
		\setlength{\tabcolsep}{6pt}
		\renewcommand{\arraystretch}{1.2}
		\begin{tabular}{@{}c p{0.75\textwidth} c c c@{}}
			\toprule
			\textbf{\#} & \textbf{Sentence} & \textbf{GT} & \textbf{SF} & \textbf{SD} \\
			\midrule
			1 & Mohamed Elshinawy, 30, is being held on a number of charges including trying to provide material support to a foreign terrorist organisation. & H & H & H \\
			2 & He was arrested in Maryland after allegedly receiving nearly \$9,000 from Islamic State (IS) operatives overseas. & \textbf{A} & H$^{\times}$ & H$^{\times}$ \\
			3 & Prosecutors say he pledged allegiance to IS and planned to use the funds for terrorist activities within the United States. & \textbf{A} & A & H$^{\times}$ \\
			4 & The FBI said Mr Elshinawy used disposable mobile phones and multiple email and social media accounts to communicate with IS contacts. & H & A$^{\times}$ & H \\
			5 & He received the money through a PayPal account and a Western Union wire transfer, the FBI said. & H & A$^{\times}$ & A$^{\times}$ \\
			6 & Those overseas financial transfers attracted the attention of authorities in June. & H & A$^{\times}$ & A$^{\times}$ \\
			7 & ``Elshinawy stated that he was instructed to use the monies he received from the unidentified ISIL operative for `operational purposes,' which Elshinawy understood to mean causing destruction or conducting a terrorist attack in the United States,'' according to an FBI affidavit. & H & H & H \\
			8 & In the past year, more than 70 people in the US have been charged with working with IS militants. & H & H & H \\
			9 & Authorities are on heightened alert this month after a California couple, who had pledged allegiance to IS, carried out the deadliest terrorist attack since the 9/11 attacks in 2001. & H & H & H \\
			10 & The FBI says it has disrupted multiple IS-linked plots this year through increased surveillance and tip-offs from the Muslim community. & \textbf{A} & A & H$^{\times}$ \\
			\bottomrule
		\end{tabular}
		\caption{Case B: a hybrid news document from XSum + Kimi. Notation follows Table~\ref{tab:case_a}. SenFlow detects an AI block but misaligns its boundaries, predicting sentences 3--6 instead of the true block at sentences 2--3, while correctly identifying the isolated AI sentence at position 10. SenDetEX misses all three AI sentences and produces two false positives.}
		\label{tab:case_b}
	\end{table*}
	
	Two pieces of internal evidence indicate it does not. First, SenFlow's margin is comparable in both transfer directions: $+4.7$ pp Macro-F1 averaged over Pub$\to$XSum, where tokenization shifts from lowercase to mixed-case, and $+3.6$ pp averaged over XSum$\to$Pub, where it shifts in the opposite direction. A tokenization-driven margin would more naturally produce a one-sided pattern in which one tokenization is the bottleneck. Second, the cross-generator protocol holds tokenization fixed within each domain: PD$\leftrightarrow$PK share PubMed formatting and XD$\leftrightarrow$XK share XSum formatting. The $+1.38$ pp Avg F1 margin observed there (Table~\ref{tab:cross_gen}) is therefore tokenization-free by construction. This provides a lower bound on SenFlow's tokenization-orthogonal advantage; the additional gain at cross-domain reflects what inter-sentence flow contributes when the domain itself shifts.
	
	A related observation concerns the baselines. SenDetEX loses its large lead over POGER and SeqXGPT on three of the four cross-domain directions (Pub$\to$XD, Pub$\to$XK, XS$\to$PD), retaining its Unified-protocol ordering only on XS$\to$PK, the direction on which both simpler baselines degrade most. SenFlow, in contrast, leads on all four directions, indicating that its inter-sentence structure transfers more reliably across domains than SenDetEX's style-context fusion.

	\section{Ablation Per-Subset Analysis}
	\label{app:ablation_full}
	
	The averaged ablation results in Table~\ref{tab:ablation} can be decomposed by subset. Removing the Hybrid Adjacency GCN consistently causes the largest Macro-F1 drop on every individual subset, with per-subset decreases of approximately 2.4, 2.1, 1.5, and 1.3 pp on PD, PK, XD, and XK respectively. The CRF, TCN, and contrastive-loss components each induce smaller drops in the 0.4--1.3 pp range, with their relative ordering varying mildly across subsets but never approaching the magnitude of the GCN drop. This confirms that the dominance of inter-sentence flow modeling is not an averaging artifact but holds across domains and generators.
	
	\paragraph{Error decomposition by sentence position.} We partition each test sentence by its ground-truth context into H-interior, A-interior, H-after-A, and A-after-H, where the suffix denotes the preceding label. Aggregated across the four MOSAIC test subsets, SenFlow and SenDetEX perform comparably on H-interior (2.85\% vs.\ 3.16\% error), but the gap widens on A-interior ($-4.49$ pp) and on A-after-H, the hardest category ($-5.70$ pp). SenFlow's per-document transition count averages 3.53, closer to the ground-truth 3.36 than SenDetEX's 3.80, indicating that the CRF decoder pulls predicted label sequences toward the block structure of MOSAIC.
	
	\section{Qualitative Case Study}
	\label{app:casestudy}
	
	To complement the quantitative comparison in \S\ref{sec:main_results}, we present two qualitative examples illustrating where inter-sentence flow differs from per-sentence classification. Both are drawn from XSum, where news sentences are shorter and easier to inspect than biomedical abstracts.

	\textbf{Case A: Inter-sentence flow recovers per-sentence errors.} Table~\ref{tab:case_a} shows a hybrid news document from the XSum + DeepSeek subset in which three sentences are AI-generated and report generic procedural details of a court hearing (sentences 5, 8, 9). SenFlow classifies all ten sentences correctly. SenDetEX makes three errors: two false positives on sentences 2 and 6, both Human but containing brief generic clauses, and one false negative on sentence 9, an AI sentence lexically adjacent to the preceding Geddes-related Human context. This illustrates a recurring failure mode of per-sentence classification: short or generically phrased sentences carry insufficient stylistic signal in isolation. Inter-sentence flow leverages neighboring context to disambiguate such cases.

	\textbf{Case B: Failure mode of label coherence on isolated AI sentences.} Table~\ref{tab:case_b} shows a hybrid news document from the XSum + Kimi subset where the AI insertions form a two-sentence block at positions 2--3 and an isolated sentence at position 10. SenFlow detects an AI block but misaligns its boundaries, predicting sentences 3--6 instead of 2--3, while still correctly flagging the isolated sentence 10. SenDetEX misses all three AI sentences. SenFlow reaches 60\% per-sentence accuracy and SenDetEX 50\%, with qualitatively different failure modes: SenDetEX produces no positive detection of the AI block, while SenFlow's errors cluster at block boundaries. This is consistent with the CRF's preference for label coherence (\S\ref{sec:flow}): once a coherent block decision is made in the boundary region it is over-extended, yet isolated AI sentences far from the boundary remain detectable.
	
\end{document}